\documentclass[journal]{IEEEtran}
\usepackage[pagebackref=true,breaklinks=true,colorlinks,citecolor=blue,linkcolor=blue,bookmarks=false]{hyperref}
\usepackage{amsmath, amssymb, amsfonts, amsthm, fouriernc, mathtools, bm,times}
\usepackage{tikz}
\usetikzlibrary{shapes,arrows,shadows}
\tikzstyle{arrow} = [thick,->,>=stealth]
\usepackage{xifthen}

\usepackage{graphicx}
\usepackage{epsfig}
\usepackage{epstopdf}
\newcommand{\norm}[1]{\left\lVert#1\right\rVert}

\include{macro}
\usepackage{array, booktabs, makecell}
\usepackage{flushend}

\usepackage{algorithmicx}
\usepackage{algorithm}
\usepackage{algpseudocode}
\usepackage{xcolor}

\usepackage[misc]{ifsym} 

\usepackage[mathscr]{euscript}
\usepackage[switch]{lineno}
\linenumbers

\pgfdeclarelayer{background}
\pgfdeclarelayer{foreground}
\pgfsetlayers{background,main,foreground}

\tikzstyle{sensor}=[draw, fill=red!20, text width=8em, 
    text centered, minimum height=0.7em,drop shadow]
\tikzstyle{agent} = [sensor, text width=8em, fill=red!50, 
    minimum height=4em, rounded corners, drop shadow]
\tikzstyle{challenge} = [sensor, text width=11em, fill=teal!40, 
    minimum height=4em, rounded corners, drop shadow]
\tikzstyle{policy} = [sensor, text width=6.25em, fill=green!20, 
    minimum height=3em, rounded corners, drop shadow]
\tikzstyle{env} = [sensor, text width=12em, fill=gray!30, 
    minimum height=6em, rounded corners, drop shadow]
\tikzstyle{env2} = [sensor, text width=6em, fill=yellow!30, 
minimum height=2.25em, rounded corners, drop shadow]
\ifCLASSOPTIONcompsoc
\usepackage[nocompress]{cite}
\else
\usepackage{cite}
\fi

\ifCLASSINFOpdf
\else
\fi
\begin{document}
\title{Deep Reinforcement Learning\\ for Autonomous Driving: A Survey}
\author{B Ravi Kiran$^1$, Ibrahim Sobh$^2$, Victor Talpaert$^3$, Patrick Mannion$^4$, \\ 
Ahmad A. Al Sallab$^2$, Senthil Yogamani$^5$, Patrick P\'erez$^6$
\thanks{$^1$Navya, Paris. \Letter\ ravi.kiran@navya.tech}
\thanks{$^2$Valeo Cairo AI team, Egypt. \Letter\ ibrahim.sobh, ahmad.el-sallab@\{valeo.com\}}
\thanks{$^3$U2IS, ENSTA Paris, Institut Polytechnique de Paris \& AKKA Technologies, France. \Letter\  victor.talpaert@ensta.fr}%
\thanks{$^4$School of Computer Science, National University of Ireland, Galway. \Letter\ patrick.mannion@nuigalway.ie}
\thanks{$^5$Valeo Vision Systems. \Letter\  senthil.yogamani@valeo.com}
\thanks{$^6$Valeo.ai. \Letter\  patrick.perez@valeo.com}
}

\nolinenumbers

\markboth{}%
{Shell \MakeLowercase{\textit{et al.}}: Bare Demo of IEEEtran.cls for Computer Society Journals}
%



\IEEEtitleabstractindextext{%
\begin{abstract}
With the development of deep representation learning, the domain of reinforcement learning (RL) has become a powerful learning framework now capable of learning complex policies in high dimensional environments. This review summarises deep reinforcement learning (DRL) algorithms and provides a taxonomy of automated driving tasks where (D)RL methods have been employed, while addressing key computational challenges in real world deployment of autonomous driving agents. It also delineates adjacent domains such as behavior cloning, imitation learning, inverse reinforcement learning that are related but are not classical RL algorithms. The role of simulators in training agents, methods to validate, test and robustify existing solutions in RL are discussed.
\end{abstract}

}

\maketitle

\IEEEdisplaynontitleabstractindextext

%
\IEEEpeerreviewmaketitle

\section{Introduction}

Autonomous driving (AD)\footnote{For easy reference, the main acronyms used in this article are listed in Appendix (Table \ref{tab:acronym1}).} systems constitute of 
multiple perception level tasks that have now achieved high precision on account of deep 
learning architectures. Besides the perception, autonomous driving systems constitute of 
multiple tasks where classical supervised learning methods are no more applicable. First, when 
the prediction of the agent's action changes future sensor observations received from the 
environment under which the autonomous driving agent operates, for example the task of optimal 
driving speed in an urban area. Second, supervisory signals such as time to collision (TTC), lateral error w.r.t to optimal trajectory of the agent, represent the dynamics of the agent, as well uncertainty in the environment. Such problems would require defining the stochastic cost function to be maximized. Third, the agent is required to learn new configurations of the environment, as well as to predict an optimal decision at each instant while driving in its 
environment. This represents a high dimensional space given the number of unique configurations
under which the agent \& environment are observed, this is combinatorially large. In all such 
scenarios we are aiming to solve a sequential decision process, which is formalized under the 
classical settings of Reinforcement Learning (RL), where the agent is required to learn and 
represent its environment as well as act optimally given at each instant \cite{sutton2018book}. The optimal action is referred to as the policy.

\begin{figure*}[ht]
    \centering
    \includegraphics[width=0.32\linewidth]{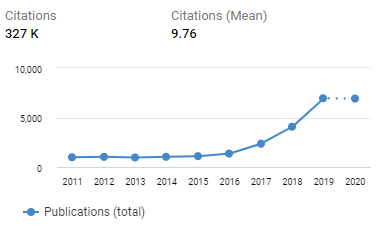}
    \includegraphics[width=0.31\linewidth]{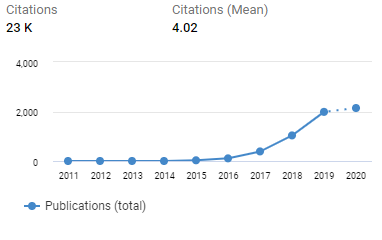}
    \includegraphics[width=0.32\linewidth]{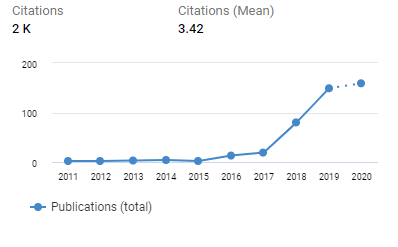}
    \caption{Trend of publications for keywords 1. "reinforcement learning", 2."deep reinforcement", and 3."reinforcement learning" AND ("autonomous cars" OR "autonomous vehicles"  OR "self driving") for academic publication trends from this \cite{dimensions_acad_trends}.}
    \label{fig:trends_rl_drl_all}
\end{figure*}

In this review we cover the notions of reinforcement learning, the taxonomy of tasks where RL is a promising solution especially in the domains of driving policy, predictive perception, path and motion planning, and low level controller design. We also focus our review on the different real world deployments of RL in the domain of autonomous driving expanding our conference paper \cite{drlvisapp19} since their deployment has not been reviewed in an academic setting. Finally, we motivate users by demonstrating the key computational challenges and risks when applying current day RL algorithms such imitation learning, deep Q learning, among others. 
We also note from the trends of publications in figure \ref{fig:trends_rl_drl_all}
that the use of RL or Deep RL applied to autonomous driving or the self driving domain is an emergent field. This is due to the recent usage of RL/DRL algorithms domain, leaving open multiple real world challenges in implementation and deployment. We address the open problems in \ref{sec:challenges}.

The main contributions of this work can be summarized as follows:
\begin{itemize}
  \item Self-contained overview of RL background for the automotive community as it is not well known. 
  \item Detailed literature review of using RL for different autonomous driving tasks.
  \item Discussion of the key challenges and opportunities for RL applied to real world  autonomous driving.  
\end{itemize}
The rest of the paper is organized as follows. Section \ref{sec:ADModules} provides an overview of components of a typical autonomous driving system.  Section \ref{sec:RL_intro} provides an introduction to reinforcement learning and briefly discusses key concepts. Section \ref{sec:RL_extensions} discusses more sophisticated extensions on top of the basic RL framework. Section \ref{sec:RL4AD} provides an overview of RL applications for autonomous driving problems. Section \ref{sec:challenges} discusses challenges in deploying RL for real-world autonomous driving systems. Section \ref{sec:conc} concludes this paper with some final remarks. 

\section{Components of AD System} \label{sec:ADModules}

\begin{figure*}[t]
\begin{center}
\includegraphics[width=0.9\linewidth]{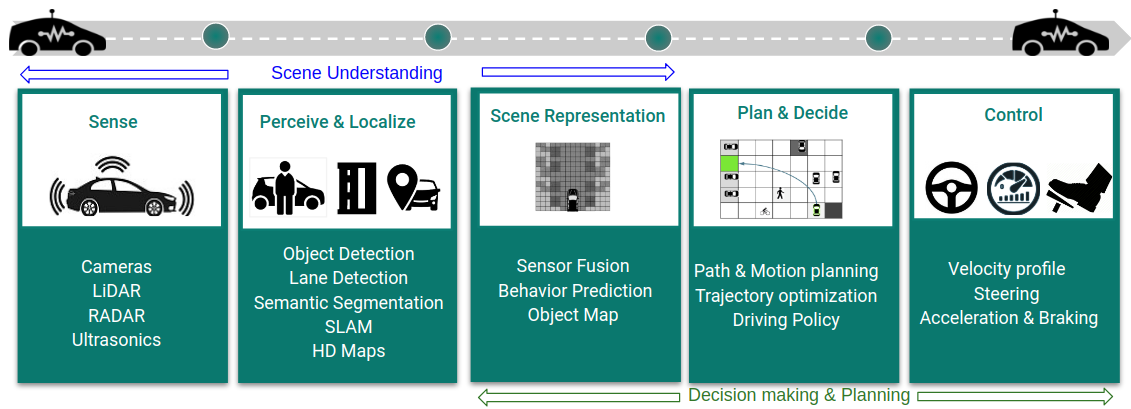}
\caption{Standard components in a modern autonomous driving systems pipeline listing the various tasks. The key problems addressed by these modules are Scene Understanding, Decision and Planning.}
\label{fig:ADflow}
\end{center}
\end{figure*}

Figure \ref{fig:ADflow} comprises of the standard blocks of an AD system demonstrating the pipeline from sensor 
stream to control actuation. The sensor architecture in a modern autonomous driving system notably includes multiple
sets of cameras, radars and LIDARs as well as a GPS-GNSS system for absolute localisation and inertial measurement Units (IMUs)
that provide 3D pose of the vehicle in space. 

The goal of the perception module is the creation of an intermediate level representation of the environment state (for example bird-eye view map of all obstacles and agents) that is to be later utilised by  a decision making system that
ultimately produces the driving policy. This state would include lane position, drivable zone, location of agents such
as cars \& pedestrians, state of traffic lights and others. Uncertainties in the perception propagate to the rest of the
information chain. Robust sensing is critical for safety thus using redundant sources increases confidence in detection. This is achieved by a combination of several perception tasks like semantic segmentation \cite{siam2017deep,el2019rgb}, motion estimation \cite{siam2018modnet}, depth estimation \cite{kumar2018monocular}, soiling detection \cite{uvrivcavr2019soilingnet}, etc which can be efficiently unified into a multi-task model \cite{sistu2019neurall,yogamani2019woodscape}.

\subsection{Scene Understanding}

This key module maps the abstract mid-level representation of the perception state obtained 
from the perception module to the high level action or decision making module. 
Conceptually, three tasks are grouped by this module:  Scene understanding, Decision and
Planning as seen in figure \ref{fig:ADflow} module aims to provide a higher level
understanding of the scene, it is built on top of the algorithmic tasks of detection or
localisation. By fusing heterogeneous sensor sources, it aims to robustly generalise to
situations as the content becomes more abstract. This information fusion provides a general
and simplified context for the Decision making components. 

Fusion provides a sensor agnostic representation of the environment and models the sensor noise and detection uncertainties across multiple modalities such as LIDAR, camera, radar, ultra-sound. This basically requires weighting the predictions in a principled way.

\subsection{Localization and Mapping}

Mapping is one of the key pillars of automated driving \cite{milz2018visual}. Once an area is mapped, current position of the vehicle can be localized within the map. 
The first reliable demonstrations of automated driving by Google
were primarily reliant on localisation to pre-mapped areas. Because of the scale of the problem, traditional mapping
techniques are augmented by semantic object detection for reliable disambiguation. In addition, localised high
definition maps (HD maps) can be used as a prior for object detection.

\subsection{Planning and Driving policy} 
Trajectory planning is a crucial module in the autonomous driving pipeline. Given
a route-level plan from HD maps or GPS based maps,
this module is required to generate motion-level commands that steer the agent.

Classical motion planning ignores dynamics and differential constraints while using
translations and rotations required to move an agent from source to destination poses
\cite{LaValle2006Book}. A robotic agent capable of controlling 6-degrees of freedom (DOF) is said to be holonomic, while an agent with fewer controllable DOFs than its total DOF is said to be non-holonomic. Classical algorithms such as $A^\ast$ algorithm based on Djisktra's algorithm do not work in the non-holonomic case for autonomous driving. Rapidly-exploring random trees (RRT) \cite{RRT2001} are non-holonomic algorithms that explore the configuration space by random sampling and obstacle free path generation. There are various versions of RRT currently used in for motion planning in autonomous driving pipelines.

\subsection{Control}

A controller defines the speed, steering angle and braking actions necessary over every
point in the path obtained from a pre-determined map such as Google maps, or expert
driving recording of the same values at every waypoint. Trajectory tracking in contrast
involves a temporal model of the dynamics of the vehicle viewing the waypoints
sequentially over time. 

Current vehicle control methods are founded in classical optimal control theory which can
be stated as a minimisation of a cost function $\dot{x} = f(x(t),u(t))$ defined over a set
of states $x(t)$ and control actions $u(t)$. The control input is usually defined over a
finite time horizon and restricted on a feasible state space $x\in X_\text{free}$
\cite{kuwata2009real}. The velocity control are based on classical methods of closed loop control such as PID (proportional-integral-derivative) controllers, MPC (Model
predictive control). PIDs aim to minimise a cost function constituting of three terms current error with proportional term, effect of past errors with integral term, and effect of future errors with the derivative term. While the family of MPC methods aim to stabilize the behavior of the vehicle while tracking the specified path \cite{paden2016survey}. A review on controllers, motion planning and learning based approaches for the same are provided in this review \cite{schwarting2018planning} for interested readers. Optimal control and reinforcement learning are intimately related, where optimal control can be viewed as a model based reinforcement learning problem where the dynamics of the vehicle/environment are modeled by well defined differential equations. Reinforcement learning methods were developed to handle stochastic control problems as well ill-posed problems with unknown rewards and state transition probabilities.
Autonomous vehicle stochastic control is large domain, and we advise readers to read the survey on this subject by authors in \cite{kuutti2020survey}.

\section{Reinforcement learning}  
\label{sec:RL_intro}
Machine learning (ML) is a process whereby a computer program learns from experience
to improve its performance at a specified task \cite{Mitchell1997Machine}. ML
algorithms are often classified under one of three broad categories: supervised
learning, unsupervised learning and reinforcement learning (RL). Supervised learning
algorithms are based on inductive inference where the model is typically trained
using labelled data to perform classification or regression, whereas unsupervised
learning encompasses techniques such as density estimation or clustering applied to
unlabelled data.
By contrast, in the RL paradigm an autonomous agent learns to improve its performance at an assigned task by interacting with its environment. Russel and Norvig define an agent as ``anything that can be viewed as perceiving its environment through sensors and acting upon that environment through actuators'' \cite{Russell2009Artificial}. RL agents are not told explicitly how to act by an expert; rather an agent's performance is evaluated by a reward function $R$. For each state experienced, the agent chooses an action and receives an occasional reward from its environment based on the usefulness of its decision. The goal for the agent is to maximize the cumulative rewards received over its lifetime. Gradually, the agent can increase its long-term reward by exploiting knowledge learned about the expected utility (i.e. discounted sum of expected future rewards) of different state-action pairs.
One of the main challenges in reinforcement learning is managing the trade-off between exploration and exploitation. To maximize the rewards it receives, an agent must exploit its knowledge by selecting actions which are known to result in high rewards. On the other hand, to discover such beneficial actions, it has to take the risk of trying new actions which may lead to higher rewards than the current best-valued actions for each system state. In other words, the learning agent has to exploit what it already knows in order to obtain rewards, but it also has to explore the unknown in order to make better action selections in the future.
Examples of strategies which have been proposed to manage this trade-off include $\epsilon$-greedy and softmax. When adopting the ubiquitous $\epsilon$-greedy strategy, an agent either selects an action at random with probability $0 < \epsilon < 1$, or greedily selects the highest valued action for the current state with the remaining probability $1 - \epsilon$. Intuitively, the agent should explore more at the beginning of the training process when little is known about the problem environment. As training progresses, the agent may gradually conduct more exploitation than exploration. The design of exploration strategies for RL agents is an area of active research (see \textit{e.g.} \cite{hong2018diversity}).

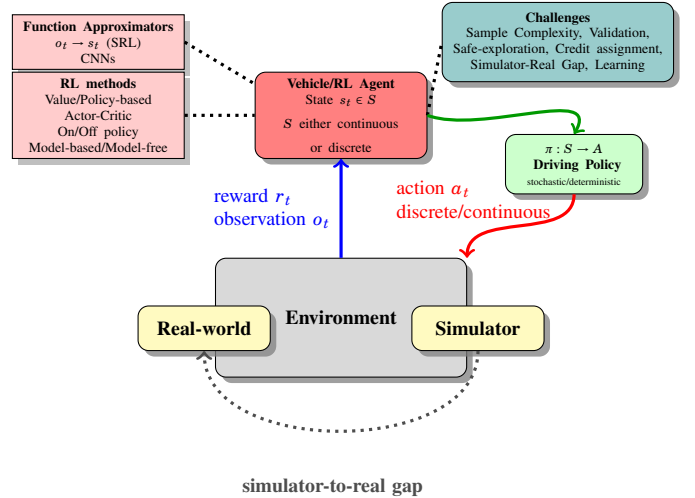
\begin{figure}
    \centering
    \begin{tikzpicture}[scale=0.65, every node/.style={scale=0.75}]
    \node (wa) [agent]  {\scriptsize {\bf Vehicle/RL Agent} \\ State $s_t \in S$\\ $S$ either continuous or discrete};
    \path (wa.west)+(+6.2,1.5) node (challenge) [challenge] {\scriptsize {\bf Challenges}\\ Sample Complexity, Validation, \\ Safe-exploration, Credit assignment,\\Simulator-Real Gap, Learning \\ };
    \path (wa.west)+(-3.2,1.5) node (func) [sensor] {\scriptsize {\bf Function Approximators}\\ $o_t \to s_t$ (SRL)\\CNNs\\};
    \path (wa.west)+(-3.2,0.0) node (rlmethod) [sensor] {\scriptsize {\bf RL methods}\\
                                                     {Value/Policy-based}\\
                                                     {Actor-Critic}\\
                                                     {On/Off policy}\\
                                                     {Model-based/Model-free}\\};
   
    \path (wa.east)+(3.0,-1.0) node (policy) [policy] {\scriptsize $\pi : S \to A$\\\ \textbf {Driving Policy}\\{\tiny stochastic/deterministic}\\};
    
    \path [draw, dotted, very thick] (func.east) -- node [above] {} (wa.160);
    \path [draw, dotted, very thick] (rlmethod.east) -- node [above] {} (wa.180);
    \path [draw, dotted, very thick] (wa.east) -- node [above] {} (challenge.180);
    \path [draw, ->, black!40!green, very thick] (wa.east) to[out=-20,in=-300] node [above] {} (policy.north);

    \path (wa.south)+(0,-3.25) node (env) [env] {\textbf{Environment}};
    \path (wa.north)+(-2.8,-5.25) node (real) [env2] {\textbf{Real-world}};
    \path (wa.north)+(+2.8,-5.25) node (sim) [env2] {\textbf{Simulator}};
    \path [draw, ->, very thick, black!70, dotted] (sim.south) to[out=-90,in=270] 
    node[anchor=center, text width=3.5cm, midway, below of=env] {\bf simulator-to-real gap} (real.south);
    \path [draw, ->, very thick, red] (policy.south) to[in=90, out=270]  node[anchor=center, text width=4.4cm, midway, above] {action $a_t$\\discrete/continuous\\} (env.north east);
    \path [draw, ->, very thick, blue] (env.north) to[out=+90,in=-90]  node[text width=2.5cm, left of=env] {{reward $r_t$} \\ {observation $o_t$}} (wa.south);;
\end{tikzpicture}
    
    \caption{A graphical decomposition of the different components of an RL algorithm. It also demonstrates the different challenges encountered while training a D(RL) algorithm.}
    \label{fig:RL_agent_env}
\end{figure}



Markov decision processes (MDPs) are considered the de facto standard when formalising sequential decision making problems involving a single RL agent \cite{Wiering2012}. An MDP consists of a set $S$ of states, a set $A$ of actions, a transition function $T$ and a reward function $R$  \cite{Puterman94}, i.e. a tuple ${<S,A,T,R>}$. When in any state ${s \in S}$, selecting an action ${a\in A}$ will result in the environment entering a new state ${s' \in S}$ with a transition probability ${T(s,a,s') \in (0,1)}$, and give a reward ${R(s,a)}$. This process is illustrated in Fig. \ref{fig:RL_agent_env}. The stochastic policy $\pi : S \to \mathcal{D}$ is a mapping from the state space to a probability over the set of actions, and $\pi(a|s)$ represents the probability of choosing action $a$ at state $s$. The goal is to find the optimal policy $\pi^*$, which results in the highest expected sum of discounted rewards \cite{Wiering2012}: 
\begin{equation}
   \pi^{*} = \underset{\pi}{\operatorname{argmax}}  
   \underbrace{\mathbb{E}_\pi \Bigg \{\sum_{k=0}^{H-1} \gamma^k r_{k+1} \mid s_0 = s \Bigg \}}_{:= V_{\pi}(s)},
    \label{eqn:optimalpolicy}
\end{equation}
for all states $s\in S$, where $r_k = R(s_k,a_k)$ is the reward at time $k$ and $V_{\pi}(s)$, the `value function' at state $s$ following a policy $\pi$, is the expected `return' (or `utility') when starting at $s$ and following the policy $\pi$ thereafter \cite{sutton2018book}. An important, related concept is the action-value function, a.k.a.`Q-function' defined as:
\begin{equation}
     Q_{\pi}(s,a) = \mathbb{E}_\pi \Bigg \{\sum_{k=0}^{H-1} \gamma^k r_{k+1} \mid s_0 = s, a_0 = a \Bigg \}.
     \label{eq:q-value}
\end{equation}
The discount factor $\gamma~\in~[0,1]$ controls how an agent regards future rewards. Low values of $\gamma$ encourage myopic behaviour where an agent will aim to maximise short term rewards, whereas high values of $\gamma$ cause agents to be more forward-looking and to maximise rewards over a longer time frame. The horizon $H$ refers to the number of time steps in the MDP. In infinite-horizon problems ${H=\infty}$, whereas in episodic domains $H$ has a finite value. Episodic domains may terminate after a fixed number of time steps, or when an agent reaches a specified goal state. The last state reached in an episodic domain is referred to as the terminal state. In finite-horizon or goal-oriented domains discount factors of (close to) 1 may be used to encourage agents to focus on achieving the goal, whereas in infinite-horizon domains lower discount factors may be used to strike a balance between short- and long-term rewards.
If the optimal policy for a MDP is known, then $V_{\pi^*}$ may be used to determine the maximum expected discounted sum of rewards available from any arbitrary initial state. A rollout is a trajectory produced in the state space by sequentially applying a policy to an initial state.
A MDP satisfies the Markov property, i.e. system state transitions are dependent only on the most recent state and action, not on the full history of states and actions in the decision process.  Moreover, in many real-world application domains, it is not possible for an agent to observe all features of the environment state; in such cases the decision-making problem is formulated as a partially-observable Markov decision process (POMDP).
Solving a reinforcement learning task means finding a policy $\pi$ that maximises the expected discounted sum of rewards over trajectories in the state space. RL agents may learn value function estimates, policies and/or environment models directly. Dynamic programming (DP) refers to a collection of algorithms that can be used to compute optimal policies given a perfect model of the environment in terms of reward and transition functions. Unlike DP, in Monte Carlo methods there is no assumption of complete environment knowledge. Monte Carlo methods are incremental in an episode-by-episode sense. Upon the completion of an episode, the value estimates and policies are updated. Temporal Difference (TD) methods, on the other hand, are incremental in a step-by-step sense, making them applicable to non-episodic scenarios. Like Monte Carlo methods, TD methods can learn directly from raw experience without a model of the environment's dynamics. Like DP, TD methods learn their estimates based on other estimates.

\subsection{Value-based methods}
\label{sec:DRL_value}
Q-learning is one of the most commonly used RL algorithms. It is a model-free TD algorithm that learns estimates of the utility of individual state-action pairs (Q-functions defined in Eqn.~\ref{eq:q-value}). 
Q-learning has been shown to converge to the optimum state-action values for a MDP with probability $1$, so long as all actions in all states are sampled infinitely often and the state-action values are represented discretely \cite{Watkins92}. In practice, Q-learning will learn (near) optimal state-action values provided a sufficient number of samples are obtained for each state-action pair. If a Q-learning agent has converged to the optimal Q values for a MDP and selects actions greedily thereafter, it will receive the same expected sum of discounted rewards as calculated by the value function with $\pi^{*}$ (assuming that the same arbitrary initial starting state is used for both). Agents implementing Q-learning update their Q values according to the following update rule:
\begin{equation}
  Q(s,a)\leftarrow Q(s,a)+\alpha[r+\gamma \max_{a'\in A}Q(s',a')-Q(s,a)],
  \label{eqn:Q-learning}
\end{equation}

\noindent where $Q(s,a)$ is an estimate of the utility of selecting action $a$ in state $s$, $\alpha~\in~[0,1]$ is the learning rate which controls the degree to which Q values are updated at each time step, and $\gamma~\in~[0,1]$ is the same discount factor used in Eqn. \ref{eqn:optimalpolicy}.
The theoretical guarantees of Q-learning hold with any arbitrary initial Q values \cite{Watkins92}; therefore the optimal Q values for a MDP can be learned by starting with any initial action value function estimate. The initialisation can be optimistic (each $Q(s,a)$ returns the maximum possible reward), pessimistic (minimum) or even using knowledge of the problem to ensure faster convergence.
Deep Q-Networks (DQN) \cite{mnih2015human} incorporates a variant of the Q-learning algorithm \cite{watkins1989learning}, by using deep neural networks (DNNs) as a non-linear Q function approximator over high-dimensional state spaces (e.g. the pixels in a frame of an Atari game). Practically, the neural network predicts the value of all actions 
without the use of any explicit domain-specific information or hand-designed features. DQN applies experience replay technique to break the correlation between successive experience samples and also for better sample efficiency. For increased stability, two networks are used where the parameters of the target network for DQN are fixed for a number of iterations while updating the parameters of the online network. Readers are directed to sub-section \ref{sec:DRL} for a more detailed introduction to the use of DNNs in Deep RL.

\subsection{Policy-based methods}
\label{sec:DRL_policy}
The difference between value-based and policy-based methods is essentially a matter of where the burden of optimality resides. Both method types must propose actions and evaluate the resulting behaviour, but while value-based methods focus on evaluating the optimal cumulative reward and have a policy follows the recommendations, policy-based methods aim to estimate the optimal policy directly, and the value is a secondary if calculated at all.
Typically, a policy is parameterised as a neural network $\pi_{\theta}$. Policy gradient methods use gradient descent to estimate the parameters of the policy that maximise the expected reward. The result can be a stochastic policy where actions are selected by sampling, or a deterministic policy. 
Many real-world applications have continuous action spaces. Deterministic policy gradient (DPG) algorithms \cite{silver2014deterministic} \cite{sutton2018book} allow reinforcement learning in domains with continuous actions. Silver \textit{et al.} \cite{silver2014deterministic} proved that a deterministic policy gradient exists for MDPs satisfying certain conditions, and that deterministic policy gradients have a simple model-free form that follows the gradient of the action-value function. As a result, instead of integrating over both state and action spaces in stochastic policy gradients, DPG integrates over the state space only leading to fewer samples in problems with large action spaces. To ensure sufficient exploration, actions are chosen using a stochastic policy, while learning a deterministic target policy.
The REINFORCE \cite{Williams1992SimpleSG} algorithm is a straight forward policy-based method. The discounted cumulative reward $g_t = \sum_{k=0}^{H-1}\gamma^{k} r_{k+t+1}$ at one time step is calculated by playing the entire episode, so no estimator is required for policy evaluation. The parameters are updated into the direction of the performance gradient:
\begin{equation}
    \theta \gets \theta + \alpha \gamma^t g \nabla \log \pi_\theta (a | s),
    \label{eqn:reinforce}
\end{equation}
where $\alpha$ is the learning rate for a stable incremental update. Intuitively, we want to encourage state-action pairs that result in the best possible returns.
Trust Region Policy Optimization (TRPO) \cite{schulman2015trust}, works by preventing the updated policies from deviating too much from previous policies, thus reducing the chance of a bad update. TRPO optimises a surrogate objective function where the basic idea is to limit each policy gradient update as measured by the Kullback-Leibler (KL) divergence between the current and the new proposed policy. This method results in monotonic improvements in policy performance. While Proximal Policy Optimization (PPO) \cite{schulman2017proximal} proposed a clipped surrogate objective function by adding a penalty for having a too large policy change. Accordingly, PPO policy optimisation is simpler to implement, and has better sample complexity while ensuring the deviation from the previous policy is relatively small.

\subsection{Actor-critic methods}
\label{sec:DRL_AC}
Actor-critic methods are hybrid methods that combine the benefits of policy-based and value-based algorithms. The policy structure that is responsible for selecting actions is known as the `actor'. The estimated value function criticises the actions made by the actor and is known as the `critic'. After each action selection, the critic evaluates the new state to determine whether the result of the selected action was better or worse than expected. Both networks need their gradient to learn. 
Let $J(\theta) := \mathbb{E}_{\pi_\theta}\left[r\right]$ represent a policy objective function, where $\theta$ designates the parameters of a DNN. Policy gradient methods search for local maximum of $J(\theta)$. Since optimization in continuous action spaces could be costly and slow, 
the DPG (Direct Policy Gradient) algorithm represents actions as parameterised function $\mu(s | \theta^\mu)$, where $\theta^\mu$ refers to the parameters of the actor network.
Then the unbiased estimate of the policy gradient gradient step is given as:
\begin{equation}
\nabla_\theta J = - \mathbb{E}_{\pi_\theta} \Big\{(g - b) \log \pi_\theta (a | s)\Big\},
\label{eqn:actorgradient}
\end{equation}
where 
$b$ is the baseline. While using $b\equiv 0$ is the simplification that leads to the REINFORCE formulation. Williams \cite{Williams1992SimpleSG} explains a well chosen baseline can reduce variance leading to a more stable learning. The baseline, $b$ can be chosen as $V_{\pi}(s)$, $Q_{\pi}(s,a)$ or `Advantage' $A_{\pi}(s,a)$ based methods.
Deep Deterministic Policy Gradient (DDPG) \cite{lillicrap2015continuous} is a model-free, off-policy (please refer to subsection \ref{subsec:mode-policy-taxonomy} for a detailed distinction), actor-critic algorithm that can learn policies for continuous action spaces using deep neural net based function approximation, extending prior work on DPG to large and high-dimensional state-action spaces. When selecting actions, exploration is performed by adding noise to the actor policy. Like DQN, to stabilise learning a replay buffer is used to minimize data correlation. A separate actor-critic specific target network is also used. Normal Q-learning is adapted with a restricted number of discrete actions, and DDPG also needs a straightforward way to choose an action. Starting from Q-learning, we extend Eqn. \ref{eq:q-value} to define the optimal Q-value and optimal action as $Q^*$ and $a^*$.

\begin{equation}
Q^*(s, a) = \underset{\pi}{\operatorname{max}}~ Q_\pi(s,a), \ \ \ a^* = \underset{a}{\operatorname{argmax}}~ Q^*(s, a).
\label{eqn:optimalaction}
\end{equation}

In the case of Q-learning, the action is chosen according to the Q-function as in Eqn. \ref{eqn:optimalaction}. But DDPG chains the evaluation of Q after the action has already been chosen according to the policy. By correcting the Q-values towards the optimal values using the chosen action, we also update the policy towards the optimal action proposition. Thus two separate networks work at estimating $Q^*$ and $\pi^*$.

Asynchronous Advantage Actor Critic (A3C) \cite{mnih2016asynchronous} uses asynchronous gradient descent for optimization of deep neural network controllers. Deep reinforcement learning algorithms based on experience replay such as DQN and DDPG have demonstrated considerable success in difficult domains such as playing Atari games. However, experience replay uses a large amount of memory to store experience samples and requires off-policy learning algorithms. In A3C, instead of using an experience replay buffer, agents asynchronously execute on multiple parallel instances of the environment. In addition to the reducing correlation of the experiences, the parallel actor-learners have a stabilizing effect on training process. This simple setup enables a much larger spectrum of on-policy as well as off-policy reinforcement learning algorithms to be applied robustly using deep neural networks.
A3C exceeded the performance of the previous state-of-the-art at the time on the Atari domain while training for half the time on a single multi-core CPU instead of a GPU by combining several ideas. It also demonstrates how using an estimate of the value function as the previously explained baseline $b$ reduces variance and improves convergence time. By defining the \textit{advantage} as $A_\pi(a, s) = Q_{\pi}(s, a) - V_\pi(s)$, the expression of the policy gradient from Eqn. \ref{eqn:actorgradient} is rewritten as $\nabla_\theta L = - \mathbb{E}_{\pi_\theta} \{A_\pi(a, s) \log \pi_\theta (a | s)\}$. The critic is trained to minimize $\frac{1}{2}\norm{A_{\pi_{\theta}}(a, s)}^2$.
The intuition of using advantage estimates rather than just discounted returns is to allow the agent to determine not just how good its actions were, but also how much better they turned out to be than expected, leading to reduced variance and more stable training. The A3C model also demonstrated good performance in 3D environments such as labyrinth exploration. Advantage Actor Critic (A2C) is a synchronous version of the asynchronous advantage actor critic model, that waits for each agent to finish its experience before conducting an update. The performance of both A2C and A3C is comparable.  
Most greedy policies must alternate between exploration and exploitation, and good exploration visits the states where the value estimate is uncertain. This way, exploration focuses on trying to find the most \textit{uncertain}
state paths as they bring valuable information. In addition to advantage, explained earlier, some methods use the entropy as the uncertainty
quantity. Most A3C implementations include this as well. 
Two methods with common authors are energy-based policies \cite{Haarnoja:2017:RLD:3305381.3305521} and more recent and with widespread use, the Soft Actor Critic (SAC) algorithm \cite{DBLP:journals/corr/abs-1812-05905}, both rely on adding an entropy term to the reward function, so we update the policy objective from Eqn. \ref{eqn:optimalpolicy} to Eqn. \ref{eqn:rewardentropy}. We refer readers to \cite{DBLP:journals/corr/abs-1812-05905} for an in depth explanation of the expression
\begin{equation}
   \pi^{*}_{MaxEnt} = \underset{\pi}{\operatorname{argmax}} \mathbb{E}_{\pi} \big\{
   \sum_t  [r(s_t, a_t) + \alpha H(\pi(. | s_t))] \big\},
    \label{eqn:rewardentropy}
\end{equation}
shown here for illustration of how the entropy $H$ is added.

\subsection{Model-based (vs. Model-free) \& On/Off Policy methods}
\label{subsec:mode-policy-taxonomy}
In practical situations, interacting with the real environment could be limited due to many reasons including safety and cost. Learning a model for environment dynamics may reduce the amount of interactions required with the real environment. Moreover, exploration can be performed on the learned models.
In the case of model-based approaches (\textit{e.g.} Dyna-Q \cite{sutton1990integrated}, R-max \cite{brafman2002r}), agents attempt to learn the transition function $T$ and reward function $R$, which can be used when making action selections. Keeping a model approximation of the environment means storing knowledge of its dynamics, and allows for fewer, and sometimes, costly environment interactions. By contrast, in model-free approaches such knowledge is not a requirement. Instead, model-free learners sample the underlying MDP directly in order to gain knowledge about the unknown model, in the form of value function estimates for example. In Dyna-2 \cite{silver2008sample}, the learning agent stores long-term and short-term memories, where a memory is defined as the set of features and corresponding parameters used by an agent to estimate the value function. Long-term memory is for general domain knowledge which is updated from real experience, while short-term memory is for specific local knowledge about the current situation, and the value function is a linear combination of long and short term memories.

Learning algorithms can be on-policy or off-policy depending on whether the updates are conducted on fresh trajectories generated by the policy or by another policy, that could be generated by an older version of the policy or provided by an expert. On-policy methods such as SARSA \cite{Rummery1994SARSA}, estimate the value of a policy while using the same policy for control. However, off-policy methods such as Q-learning \cite{watkins1989learning}, use two policies: the behavior policy, the policy used to generate behavior; and the target policy, the one being improved on. An advantage of this separation is that the target policy may be deterministic (greedy), while the behavior policy can continue to sample all possible actions, \cite{sutton2018book}.

\subsection{Deep reinforcement learning (DRL)}
\label{sec:DRL}
Tabular representations are the simplest way to store learned estimates (of \textit{e.g.} values, policies or models), where each state-action pair has a discrete estimate associated with it. When estimates are represented discretely, each additional feature tracked in the state leads to an exponential growth in the number of state-action pair values that must be stored \cite{sutton2015reinforcement}. This problem is commonly referred to in the literature as the ``curse of dimensionality'', a term originally coined by Bellman \cite{Bellman1957Dynamic}. In simple environments this is rarely an issue, but it may lead to an intractable problem in real-world applications, due to memory and/or computational constraints. Learning over a large state-action space is possible, but may take an unacceptably long time to learn useful policies. Many real-world domains feature continuous state and/or action spaces; these can be discretised in many cases. However, large discretisation steps may limit the achievable performance in a domain, whereas small discretisation steps may result in a large state-action space where obtaining a sufficient number of samples for each state-action pair is impractical.
Alternatively, function approximation may be used to generalise across states and/or actions, whereby a function approximator is used to store and retrieve estimates. Function approximation is an active area of research in RL, offering a way to handle continuous state and/or action spaces, mitigate against the state-action space explosion and generalise prior experience to previously unseen state-action pairs. Tile coding is one of the simplest forms of function approximation, where one tile represents multiple states or state-action pairs \cite{sutton2015reinforcement}. 
Neural networks are also commonly used to implement function approximation, one of the most famous examples being Tesuaro's application of RL to backgammon \cite{Tesauro1994TD}. Recent work has applied deep neural networks as a function approximation method; this emerging paradigm is known as deep reinforcement learning (DRL). DRL algorithms have achieved human level performance (or above) on complex tasks such as playing Atari games \cite{mnih2015human} and playing the board game Go \cite{silver2016mastering}.

In DQN \cite{mnih2015human} it is demonstrated how a convolutional neural network can learn successful control policies from just raw video data for different Atari environments. The network was trained end-to-end and was not provided with any game specific information. The input to the convolutional neural network consists of a $84\times84\times4$ tensor of 4 consecutive stacked frames used to capture the temporal information. Through consecutive layers, the network learns how to combine features in order to identify the action most likely to bring the best outcome. One layer consists of several convolutional filters. For instance, the first layer uses 32 filters with $8\times8$ kernels with stride 4 and applies a rectifier non linearity. The second layer is 64 filters of $4\times4$ with stride 2, followed by a rectifier non-linearity. Next comes a third convolutional layer of 64 filters of $3\times3$ with stride 1 followed by a rectifier. The last intermediate layer is composed of 512 rectifier units fully connected. The output layer is a fully-connected linear layer with a single output for each valid action.
For DQN training stability, two networks are used while the parameters of the target network are fixed for a number of iterations while updating the online network parameters. For practical reasons, the $Q(s,a)$ function is modeled as a deep neural network that predicts the value of all actions given the input state. Accordingly, deciding what action to take  requires performing a single forward pass of the network.
Moreover, in order to increase sample efficiency, experiences of the agent are stored in a replay memory (experience replay), where the Q-learning updates are conducted on randomly selected samples from the replay memory. This random selection breaks the correlation between successive samples. Experience replay enables reinforcement learning agents to \textit{remember} and reuse experiences from the past where observed transitions are stored for some time, usually in a queue, and sampled uniformly from this memory to update the network. However, this approach simply replays transitions at the same frequency that they were originally experienced, regardless of their significance. An alternative method is to use two separate experience buckets, one for positive and one for negative rewards \cite{narasimhan2015language}. Then a fixed fraction from each bucket is selected to replay. This method is only applicable in domains that have a natural notion of binary experience. Experience replay has also been extended with a framework for prioritising experience \cite{schaul2015prioritized}, where important transitions, based on the TD error, are replayed more frequently, leading to improved performance and faster training when compared to the standard experience replay approach. 

The max operator in standard Q-learning and DQN uses the same values both to select and to evaluate an action resulting in over optimistic value estimates. In Double DQN (D-DQN) \cite{van2016deep} the over estimation problem in DQN is tackled where the greedy policy is evaluated according to the online network and uses the target network to estimate its value. It was shown that this algorithm not only yields more accurate value estimates, but leads to much higher scores on several games. 

In Dueling network architecture \cite{wang2015dueling} the state value function and associated advantage function are estimated, and then combined together to estimate action value function. The advantage of the dueling architecture lies partly in its ability to learn the state-value function efficiently. In a single-stream architecture only the value for one of the actions is updated. However in dueling architecture, the value stream is updated with every update, allowing for better approximation of the state values, which in turn need to be accurate for temporal difference methods like Q-learning.

DRQN \cite{hausknecht2015deep} applied a modification to the DQN by combining a Long Short Term Memory (LSTM) with a Deep Q-Network. Accordingly, the DRQN is capable of integrating information across frames to detect information such as velocity of objects. DRQN showed to generalize its policies in case of complete observations and when trained on Atari games and evaluated against flickering games, it was shown that DRQN generalizes better than DQN.  

\section{Extensions to reinforcement learning} \label{sec:RL_extensions}
This section introduces and discusses some of the main extensions to the basic single-agent RL paradigms which have been introduced over the years. As well as broadening the applicability of RL algorithms, many of the extensions discussed here have been demonstrated to improve scalability, learning speed and/or converged performance in complex problem domains.

\subsection{Reward shaping}
As noted in Section \ref{sec:RL_intro}, the design of the reward function is crucial: RL agents seek to maximise the return from the reward function, therefore the optimal policy for a domain is defined with respect to the reward function. In many real-world application domains, learning may be difficult due to sparse and/or delayed rewards. RL agents typically learn how to act in their environment guided merely by the reward signal. Additional knowledge can be provided to a learner by the addition of a shaping reward to the reward naturally received from the environment, with the goal of improving learning speed and converged performance. This principle is referred to as reward shaping. 
The term shaping has its origins in the field of experimental psychology, and describes the idea of rewarding all behaviour that leads to the desired behaviour. Skinner \cite{Skinner1938Behavior} discovered while training a rat to push a lever that any movement in the direction of the lever had to be rewarded to encourage the rat to complete the task. Analogously to the rat, a RL agent may take an unacceptably long time to discover its goal when learning from delayed rewards, and shaping offers an opportunity to speed up the learning process.
Reward shaping allows a reward function to be engineered in a way to provide more frequent feedback signal on appropriate behaviours \cite{Wiewiora2017Reward}, which is especially useful in domains with sparse rewards. Generally, the return from the reward function is modified as follows: $r' = r + f$ where $r$ is the return from the original reward function $R$, $f$ is the additional reward from a shaping function $F$, and $r'$ is the signal given to the agent by the augmented reward function $R'$. 
Empirical evidence has shown that reward shaping can be a powerful tool to improve the learning speed of RL agents \cite{Randlov98}. However, it can have unintended consequences. The implication of adding a shaping reward is that a policy which is optimal for the augmented reward function $R'$ may not in fact also be optimal for the original reward function $R$. A classic example of reward shaping gone wrong for this exact reason is reported by \cite{Randlov98} where the experimented bicycle agent would turn in circle to stay upright rather than reach its goal.
Difference rewards (\textit{D}) \cite{Wolpert2000Collective} and potential-based reward shaping (\textit{PBRS}) \cite{Ng99} are two commonly used shaping approaches. Both \textit{D} and \textit{PBRS} have been successfully applied to a wide range of application domains and have the added benefit of convenient theoretical guarantees, meaning that they do not suffer from the same issues as the unprincipled reward shaping approaches described above (see \textit{e.g.} \cite{Ng99,Devlin2011Theoretical,Mannion2017Policy,Colby2015Evolutionary,Mannion2017Theoretical}).

\subsection{Multi-agent reinforcement learning (MARL)}
\label{sec:MARL}
In multi-agent reinforcement learning, multiple RL agents are deployed into a common environment. The single-agent MDP framework becomes inadequate when multiple autonomous agents act simultaneously in the same domain. Instead, the more general stochastic game (SG) may be used in the case of a Multi-Agent System (MAS) \cite{Busoniu10}. A SG is defined as a tuple $<~S,A_{1...N},T,R_{1...N}~>$, where $N$ is the number of agents, $S$ is the set of system states, $A_i$ is the set of actions for agent $i$ (and $A$ is the joint action set), $T$ is the transition function, and $R_i$ is the reward function for agent $i$.
The SG looks very similar to the MDP framework, apart from the addition of multiple agents. In fact, for the case of $N=1$ a SG then becomes a MDP. The next system state and the rewards received by each agent depend on the joint action $a$ of all of the agents in a SG, where $a$ is derived from the combination of the individual actions $a_i$ for each agent in the system. Each agent may have its own local state perception $s_i$, which is different to the system state $s$ (i.e. individual agents are not assumed to have full observability of the system). 
Note also that each agent may receive a different reward for the same system state transition, as each agent has its own separate reward function $R_i$. In a SG, the agents may all have the same goal (collaborative SG), totally opposing goals (competitive SG), or there may be elements of collaboration and competition between agents (mixed SG). Whether RL agents in a MAS will learn to act together or at cross-purposes depends on the reward scheme used for a specific application.

\subsection{Multi-objective reinforcement learning}
In multi-objective reinforcement learning (MORL) the reward signal is a vector, where each component represents the performance on a different objective. The MORL framework was developed to handle sequential decision making problems where tradeoffs between conflicting objective functions must be considered. Examples of real-world problems with multiple objectives include selecting energy sources (tradeoffs between fuel cost and emissions) \cite{Mannion2016Multi} and watershed management (tradeoffs between generating electricity, preserving reservoir levels and supplying drinking water) \cite{Mason2016Applying}. Solutions to MORL problems are often evaluated using the concept of Pareto dominance \cite{Pareto1906Manual} and MORL algorithms typically seek to learn or approximate the set of non-dominated solutions.
MORL problems may be defined using the MDP or SG framework as appropriate, in a similar manner to single-objective problems. The main difference lies in the definition of the reward function: instead of returning a single scalar value $r$, the reward function $\mathbf{R}$ in multi-objective domains returns a vector $\mathbf{r}$ consisting of the rewards for each individual objective $c \in C$. Therefore, a regular MDP or SG can be extended to a Multi-Objective MDP (MOMDP) or Multi-Objective SG (MOSG) by modifying the return of the reward function. For a more complete overview of MORL beyond the brief summary presented in this section, the interested reader is referred to recent surveys \cite{Roijers2013Survey,ruadulescu2020multi}.

\subsection{State Representation Learning (SRL)}
State Representation Learning refers to feature extraction \& dimensionality reduction to represent the state space with its history conditioned by the actions and environment of the agent. A complete review of SRL for control is discussed in \cite{LESORT2018379}. In the simplest form SRL maps a high dimensional vector $o_t$ into a small dimensional latent space $s_t$. The inverse operation decodes the state back into an estimate of the original observation $\hat{o}_t$. The agent then learns to map from the latent space to the action. Training the SRL chain is unsupervised in the sense that no labels are required. Reducing the dimension of the input effectively simplifies the task as it removes noise and decreases the domain's size as shown in \cite{DBLP:journals/corr/abs-1901-08651}. SRL could be a simple auto-encoder (AE), though various methods exist for observation reconstruction such as Variational Auto-Encoders (VAE) or Generative Adversarial Networks (GANs), as well as forward models for predicting the next state  or inverse models for predicting the action given a transition. A good learned state representation should be Markovian; i.e. it should encode all necessary information to be able to select an action based on the current state only, and not any previous states or actions \cite{bohmer2015autonomous,LESORT2018379}.



\subsection{Learning from Demonstrations}
Learning from Demonstrations (LfD) is used by humans to acquire new skills in an expert to learner knowledge transmission process. LfD is important for initial exploration where reward signals are too sparse or the input domain is too large to cover. In LfD, an agent learns to perform a task from demonstrations, usually in the form of state-action pairs, provided by an expert without any feedback rewards. However, high quality and diverse demonstrations are hard to collect, leading to learning sub-optimal policies. Accordingly, learning merely from demonstrations can be used to initialize the learning agent with a good or safe policy, and then reinforcement learning can be conducted to enable the discovery of a better policy by interacting with the environment. Combining demonstrations and reinforcement learning has been conducted in recent research.
AlphaGo \cite{silver2016mastering}, combines search tree with deep neural networks, initializes the policy network by supervised learning on state-action pairs provided by recorded games played by human experts. Additionally, a value network is trained to tell how desirable a board state is. By conducting self-play and reinforcement learning, AlphaGo is able to discover new stronger actions and learn from its mistakes, achieving super human performance. More recently, AlphaZero \cite{silver2017mastering}, developed by the same team, proposed a general framework for self-play models. AlphaZero is trained entirely using reinforcement learning and self play, starting from completely random play, and requires no prior knowledge of human players. AlphaZero taught itself from scratch how to master the games of chess, shogi, and Go game, beating a world-champion program in each case. In \cite{abbeel2005exploration} it is shown that given the initial demonstration, no explicit exploration is necessary, and we can attain near-optimal performance. Measuring the divergence between the current policy and the expert policy for optimization is proposed in \cite{kang2018policy}. DQfD \cite{hester2018deep} pre-trains the agent and uses expert demonstrations by adding them into the replay buffer with additional priority. Moreover, a training framework that combines learning from both demonstrations and reinforcement learning is proposed in \cite{sobh2018fast} for fast learning agents. Two policies close to maximizing the reward function can still have large differences in behaviour. To avoid degenerating a solution which would fit the reward but not the original behaviour, authors \cite{abbeel2004apprenticeship} proposed a method for enforcing that the optimal policy learnt over the rewards should still match the observed policy in behavior.
Behavior Cloning (BC) is applied as a supervised learning that maps states to actions based on demonstrations provided by an expert. On the other hand, Inverse Reinforcement Learning (IRL) is about inferring the reward function that justifies demonstrations of the expert. IRL is the problem of extracting a reward function given observed, optimal behavior \cite{ng2000algorithms}. A key motivation is that the reward function provides a succinct and robust definition of a task. 
Generally, IRL algorithms can be expensive to run, requiring reinforcement learning in an inner loop between cost estimation to policy training and evaluation. Generative Adversarial Imitation Learning (GAIL) \cite{ho2016generative} introduces a way to avoid this expensive inner loop. In practice, GAIL trains a policy close enough to the expert policy to fool a discriminator. This process is similar to GANs \cite{NIPS2014_5423,uvrivcavr2019yes}. The resulting policy must travel the same MDP states as the expert, or the discriminator would pick up the differences. The theory behind GAIL is an equation simplification: qualitatively, if IRL is going from demonstrations to a cost function and RL from a cost function to a policy, then we should altogether be able to go from demonstration to policy in a single equation while avoiding the cost function estimation.


\section{Reinforcement learning for Autonomous driving tasks} 
\label{sec:RL4AD}
Autonomous driving tasks where RL could be applied include:
controller optimization, path planning and trajectory optimization, motion planning and dynamic path planning, development of high-level driving policies for complex navigation tasks, scenario-based policy learning for highways, intersections, merges and splits, reward learning with inverse reinforcement learning from expert data for intent prediction for traffic actors such as pedestrian, vehicles and finally learning of policies that ensures safety and perform risk estimation. Before discussing the applications of DRL to AD tasks we briefly review the state space, action space and rewards schemes in autonomous driving setting.

\subsection{State Spaces, Action Spaces and Rewards}
To successfully apply DRL to autonomous driving tasks, designing appropriate state spaces, action spaces, and reward functions is important. Leurent \textit{et al.} \cite{leurent2018survey} provided a comprehensive review of the different state and action representations which are used in autonomous driving research. Commonly used state space features for an autonomous vehicle include: position, heading and velocity of ego-vehicle, as well as other obstacles in the sensor view extent of the ego-vehicle. To avoid variations in the dimension of the state space, a Cartesian or Polar occupancy grid around the ego vehicle is frequently employed. This is further augmented with lane information such as lane number (ego-lane or others), path curvature, past and future trajectory of the ego-vehicle, longitudinal information such as Time-to-collision (TTC), and finally scene information such as traffic laws and signal locations. 

\begin{figure*}[t]
\begin{center}
\includegraphics[width=0.9\linewidth]{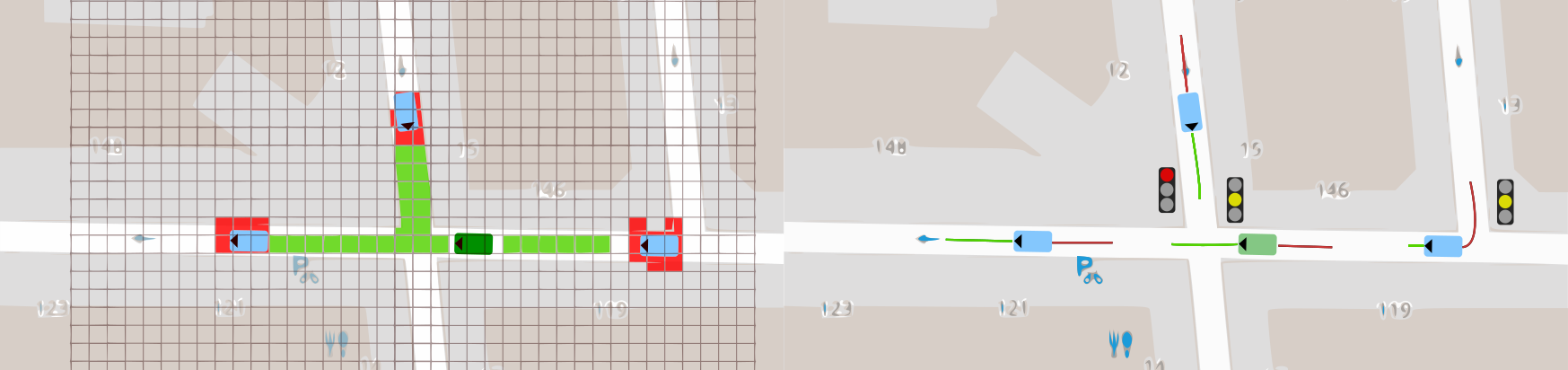}
\caption{Bird Eye View (BEV) 2D representation of a driving scene. Left demonstrates an occupancy grid. Right shows the combination of semantic information (traffic lights) with past (red) and projected (green) trajectories. The ego car is represented by a green rectangle in both images.}
\label{fig:bev}
\end{center}
\end{figure*}

Using raw sensor data such as camera images, LiDAR, radar, etc. provides the benefit of finer contextual information, while using condensed abstracted data reduces the complexity of the state space. In between, a mid-level representation such as 2D bird eye view (BEV) is sensor agnostic but still close to the spatial organization of the scene. Fig. \ref{fig:bev} is an illustration of a top down view showing an occupancy grid, past and projected trajectories, and semantic information about the scene such as the position of traffic lights. This intermediary format retains the spatial layout of roads when graph-based representations would not. Some simulators offer this view such as Carla or Flow (see Table \ref{tab:simulators}).

A vehicle policy must control a number of different actuators. Continuous-valued actuators for vehicle control include steering angle, throttle and brake. Other actuators such as gear changes are discrete. To reduce complexity and allow the application of DRL algorithms which work with discrete action spaces only (\textit{e.g.} DQN), an action space may be discretised uniformly by dividing the range of continuous actuators such as steering angle, throttle and brake into equal-sized bins (see Section \ref{sec:sample_efficiency}). Discretisation in log-space has also been suggested, as many steering angles which are selected in practice are close to the centre \cite{xu2017end}. Discretisation does have disadvantages however; it can lead to jerky or unstable trajectories if the step values between actions are too large. Furthermore, when selecting the number of bins for an actuator there is a trade-off between having enough discrete steps to allow for smooth control, and not having so many steps that action selections become prohibitively expensive to evaluate. As an alternative to discretisation, continuous values for actuators may also be handled by DRL algorithms which learn a policy directly, (\textit{e.g.} DDPG). Temporal abstractions options framework \cite{sutton1999between}) may also be employed to simplify the process of selecting actions, where agents select \textit{options} instead of low-level actions. These options represent a sub-policy that could extend a primitive action over multiple time steps.

Designing reward functions for DRL agents for autonomous driving is still very much an open question. Examples of criteria for AD tasks include: distance travelled towards a destination \cite{Dosovitskiy17}, speed of the ego vehicle \cite{Dosovitskiy17,li2019urban, kardell2017autonomous}, keeping the ego vehicle at a standstill \cite{chen2019model}, collisions with other road users or scene objects \cite{Dosovitskiy17,li2019urban}, infractions on sidewalks \cite{Dosovitskiy17}, keeping in lane, and maintaining comfort and stability  while avoiding extreme acceleration, braking or steering \cite{chen2019model,kardell2017autonomous}, and following traffic rules \cite{li2019urban}. 

\subsection{Motion Planning \& Trajectory optimization}

\begin{table*}[ht]
\centering
 \begin{tabular}{l p{7cm} p{7cm}} 
 \toprule
AD Task & (D)RL method \& description & Improvements \& Tradeoffs \\ \midrule
 
 Lane Keep & 1. Authors \cite{sallab2016end} propose a DRL system for discrete actions (DQN) and continuous actions (DDAC) using the TORCS simulator (see Table \ref{tab:simulators}) 2. Authors  \cite{sallab2017deep} learn discretised and continuous policies using DQNs and Deep Deterministic Actor Critic (DDAC) to follow the lane and maximize average velocity.& 1. This study concludes that using continuous actions provide smoother trajectories, though on the negative side lead to more restricted termination conditions \& slower convergence time to learn. 
 2. Removing memory replay in DQNs help for faster convergence \& better performance. The one hot encoding of action space resulted in abrupt steering control. While DDAC's continuous policy helps smooth the actions and provides better performance.\\  
 \hline
Lane Change &  Authors \cite{wang2018reinforcement} use Q-learning to learn a policy for ego-vehicle to perform no operation, lane change to left/right, accelerate/decelerate. & This approach is more robust compared to traditional approaches which consist in defining fixed way points, velocity profiles and curvature of path to be followed by the ego vehicle. \\ 
\hline
Ramp Merging & Authors \cite{wang2017formulation} propose recurrent architectures namely LSTMs to model longterm dependencies for ego vehicles ramp merging into a highway. & Past history of the state information is used to perform the merge more robustly.\\ 
\hline
 Overtaking & Authors \cite{ngai2011multiple} propose Multi-goal RL policy that is learnt by Q-Learning or Double action Q-Learning(DAQL) is employed
to determine individual action decisions based on whether
the other vehicle interacts with the agent for that particular
goal. & Improved speed for lane keeping and overtaking with collision avoidance.\\ 
\hline
Intersections &  Authors use DQN to evalute the Q-value for state-action pairs to negotiate intersection \cite{isele2018navigating},  & Creep-Go actions defined by authors enables the vehicle to maneuver intersections with restricted spaces and visibility more safely \\
 \hline
 Motion Planning & Authors \cite{keselman2018reinforcement} propose an improved $A^\ast$ algorithm to learn a heuristic function using deep neural netowrks over image-based input obstacle map  & Smooth control behavior of vehicle and better peformance compared to multi-step DQN\\ 
 
\bottomrule
 \end{tabular}
 \label{tab:DRLAD}
 \vspace{5pt}
 \caption{List of AD tasks that require D(RL) to learn a policy or behavior.}
\end{table*}

Motion planning is the task of ensuring the existence of a path between target and destination 
points. This is necessary to plan trajectories for vehicles over prior maps usually augmented
with semantic information. Path planning in dynamic environments and varying vehicle dynamics is a key problem in autonomous driving, for example negotiating right to pass through in an intersection \cite{isele2018navigating}, merging into highways. Recent work by authors \cite{interactiondataset} contains real world motions by various traffic actors, observed in diverse interactive driving scenarios.
Recently, authors demonstrated an application of DRL (DDPG) for AD using a full-sized autonomous vehicle \cite{kendall2019learning}. The system was first trained in simulation, before being trained in real time using on board computers, and was able to learn to follow a lane, successfully completing a real-world trial on a 250 metre section of road. 
Model-based deep RL algorithms have been proposed for learning models and policies directly from raw pixel inputs \cite{watter2015embed}, \cite{wahlstrom2015learning}. In \cite{chiappa2017recurrent}, deep neural networks have been used to generate predictions in simulated environments over hundreds of time steps.
RL is also suitable for Control. Classical optimal control methods like LQR/iLQR are compared with RL methods in \cite{recht2018tour}. Classical RL methods are used to perform optimal control in stochastic settings, for example the Linear Quadratic Regulator (LQR) in linear regimes and iterative LQR (iLQR) for non-linear regimes are utilized. A recent study in \cite{mania2018simple} demonstrates that random search over the parameters for a policy network can perform as well as LQR.

\subsection{Simulator \& Scenario generation tools}

\begin{table*}[ht]
\centering
 \begin{tabular}{l p{12cm}} \toprule
 Simulator & Description \\ \midrule 
 CARLA \cite{Dosovitskiy17} &  Urban simulator, Camera \& LIDAR streams,  
with depth \& semantic segmentation, Location information \\ 
 TORCS \cite{wymann2000torcs} & Racing Simulator, Camera stream, agent positions, testing control policies for vehicles  \\
 AIRSIM \cite{shah2018airsim} & Camera stream with depth and semantic segmentation, support for drones  \\
 GAZEBO (ROS) \cite{Koenig04designand_GAZEBO} & Multi-robot physics simulator employed for path 
planning \& vehicle control in complex 2D \& 3D maps \\
 SUMO \cite{SUMO2018} & Macro-scale modelling of traffic in cities motion planning simulators are used\\
 DeepDrive \cite{craig_quiter_2018_1248998} & Driving simulator based on unreal, providing multi-camera (eight) stream with depth \\
 Constellation \cite{constellation_nvidia} & NVIDIA DRIVE Constellation\textsuperscript{TM} simulates camera, LIDAR \& radar for AD (Proprietary)\\
 MADRaS \cite{Madras} & Multi-Agent Autonomous Driving Simulator built on top of TORCS\\
Flow \cite{DBLP:journals/corr/abs-1710-05465} & Multi-Agent Traffic Control Simulator built on top of SUMO\\
 Highway-env \cite{highway-env}  & A gym-based environment that provides a simulator for highway based road topologies \\
 Carcraft & Waymo's simulation environment (Proprietary) \\ \bottomrule
 \end{tabular}
 \label{tab:simulators}
 \vspace{5pt}
 \caption{Simulators for RL applications in advanced driving assistance systems (ADAS) and autonomous driving.}
\end{table*}

Autonomous driving datasets address supervised learning setup with training sets containing image, label pairs for various modalities. Reinforcement learning requires an environment where state-action pairs can be recovered while modelling dynamics of the vehicle state, environment as well as the stochasticity in the movement and actions of the environment and agent respectively. Various simulators are actively used for training and validating reinforcement learning algorithms. Table \ref{tab:simulators} summarises various high fidelity perception simulators capable of simulating cameras, LiDARs and radar. Some simulators are also capable of providing the vehicle state and dynamics. A complete review of sensors and simulators utilised within the autonomous driving community is available in \cite{rosique2019systematic} for readers.
Learned driving policies are stress tested in simulated environments before moving on to costly evaluations in the real world. Multi-fidelity reinforcement learning (MFRL) framework is proposed in  \cite{cutler2014reinforcement} where multiple simulators are available. In MFRL, a cascade of simulators with increasing fidelity are used in representing state dynamics (and thus computational cost) that enables the training and validation of RL algorithms, while finding near optimal policies for the real world with fewer expensive real world samples using a remote controlled car.
CARLA Challenge \cite{carla-challenge} is a Carla simulator based AD competition with pre-crash scenarios characterized in a National Highway Traffic Safety Administration report \cite{najm2007pre}. The systems are evaluated in critical scenarios such as: Ego-vehicle loses control, ego-vehicle reacts to unseen obstacle, lane change to evade slow leading vehicle among others. The scores of agents are evaluated as a function of the aggregated distance travelled in different circuits, and total points discounted due to infractions.

\subsection{LfD and IRL for AD applications}
Early work on Behavior Cloning (BC) for driving cars in \cite{pomerleau1989alvinn}, \cite{pomerleau1991efficient} presented agents that learn form demonstrations (LfD) that tries to mimic the behavior of an expert. BC is typically implemented as a supervised learning, and accordingly, it is hard for BC to adapt to new, unseen situations. An architecture for learning a convolutional neural network, end to end, in self-driving cars domain was proposed in \cite{bojarski2016end,bojarski2017explaining}. The CNN is trained to map raw pixels from a single front facing camera directly to steering commands. Using a relatively small training dataset from humans/experts, the system learns to drive in traffic on local roads with or without lane markings and on highways. The network learns image representations that detect the road successfully, without being explicitly trained to do so. 
Authors of \cite{kuderer2015learning} proposed to learn comfortable driving trajectories optimization using expert demonstration from human drivers using Maximum Entropy Inverse RL.
Authors of \cite{sharifzadeh2016learning} used DQN as the refinement step in IRL to extract the rewards, in an effort learn human-like lane change behavior. 

\section{Real world challenges and future perspectives}
In this section, challenges for conducting reinforcement learning for real-world autonomous  driving are presented and discussed along with the related research approaches for solving them.    

\label{sec:challenges}
\subsection{Validating RL systems}
Henderson \textit{et al.} \cite{henderson2018deep} described challenges in validating reinforcement learning methods focusing on policy gradient  methods for continuous control algorithms such as PPO, DDPG and TRPO as well as in reproducing benchmarks. They demonstrate with real examples that implementations often have varying code-bases and different hyper-parameter values, and that unprincipled ways to estimate the top-k rollouts could lead to incoherent interpretations on the performance of the reinforcement learning algorithms, and further more on how well they generalize. Authors concluded that evaluation could be performed either on a well defined common setup or on real-world tasks.
Authors in \cite{abeysirigoonawardena2019generating} proposed automated generation of challenging and rare driving scenarios in high-fidelity photo-realistic simulators. These adversarial scenarios are automatically discovered by parameterising the behavior of pedestrians and other vehicles on the road. Moreover, it is shown that by adding these scenarios to the training data of imitation learning, the safety is increased. 


\subsection{Bridging the simulation-reality gap} 
Simulation-to-real-world transfer learning is an active domain, since simulations are a source large \& cheap data with perfect annotations.
Authors \cite{bousmalis2017using} train a robot arm to grasp objects in the real world by performing domain adaption from simulation-to-reality, at both feature-level and pixel-level. The vision-based grasping system achieved comparable performance with 50 times fewer real-world samples.  Authors in \cite{peng2017sim}, randomized the dynamics of the simulator during training. The resulting policies were capable of generalising to different dynamics without requiring retraining on real system. In the domain of autonomous driving, authors \cite{pan2017virtual} train an A3C agent using simulation-real translated images of the driving environment. Following which, the trained policy was evaluated on a real world driving dataset. 

Authors in \cite{bewley2019learning} addressed the issue of performing imitation learning in simulation that transfers well to images from real world. They achieved this by unsupervised domain translation between simulated and real world images, that enables learning the prediction of steering in the real world domain with only ground truth from the simulated domain. Authors remark that there were no pairwise correspondences between images in the simulated training set and the unlabelled real-world image set. Similarly, \cite{Vr-goggles} performs domain adaptation to map real world images to simulated images. In contrast to sim-to-real methods they handle the reality gap during deployment of agents in real scenarios, by adapting the real camera streams to the synthetic modality, so as to map the unfamiliar or unseen features of real images back into the simulated environment and states. The agents have already learnt a policy in simulation.

\subsection{Sample efficiency} 
\label{sec:sample_efficiency}
Animals are usually able to learn new tasks in just a few trials, benefiting from their prior knowledge about the environment. However, one of the key challenges for reinforcement learning is sample efficiency. The learning process requires too many samples to learn a reasonable policy. This issue becomes more noticeable when collection of valuable experience is expensive or even risky to acquire. In the case of robot control and autonomous driving, sample efficiency is a difficult issue due to the delayed and sparse rewards found in typical settings, along with an unbalanced distribution of observations in a large state space.

\textbf{Reward shaping} enables the agent to learn intermediate goals by designing a more frequent reward function to encourage the agent to learn faster from fewer samples. Authors in \cite{Chae2017AutonomousBS} design a second "trauma" replay memory that contains only collision situations in order to pool positive and negative experiences at each training step. 

\textbf{IL boostrapped RL}: Further efficiency can be achieved where the agent first learns an initial policy offline performing imitation learning from roll-outs provided by an expert. Following which, the agent can self-improve by applying RL while interacting with the environment. 
 
 \textbf{Actor Critic} with Experience Replay (ACER) \cite{wang2016sample}, is a sample-efficient policy gradient algorithm that makes use of a replay buffer, enabling it to perform more than one gradient update using each piece of sampled experience, as well as a trust region policy optimization method.
    
\textbf{Transfer learning} is another approach for sample efficiency, which enables the reuse of previously trained policy for a source task to initialize the learning of a target task. Policy composition presented in \cite{liaw2017composing} propose composing previously learned basis policies to be able to reuse them for a novel task, which leads to faster learning of new policies. A survey on transfer learning in RL is presented in \cite{taylor2009transfer}. Multi-fidelity reinforcement learning (MFRL) framework \cite{cutler2014reinforcement} showed to transfer heuristics to guide exploration in high fidelity simulators and find near optimal policies for the real world with fewer real world samples.
Authors in \cite{isele2017transferring} transferred policies learnt to handle simulated intersections to real world examples between DQN agents.

\textbf{Meta-learning} algorithms enable agents adapt to new tasks and learn new
skills rapidly from small amounts of experiences, benefiting from their prior
knowledge about the world. Authors of \cite{wang2016learning} addressed this issue
through training a recurrent neural network on a training set of interrelated tasks,
where the network input includes the action selected in addition to the reward
received in the previous time step. Accordingly, the agent is trained to learn to
exploit the structure of the problem dynamically and solve new problems by adjusting
its hidden state. A similar approach for designing RL algorithms is presented in
\cite{duan2016rl}. Rather than designing a ``fast'' reinforcement learning
algorithm, it is represented as a recurrent neural network, and learned from data.
In Model-Agnostic Meta-Learning (MAML) proposed in \cite{finn2017model}, the
meta-learner seeks to find an initialisation for the parameters of a neural network,
that can be adapted quickly for a new task using only few examples. Reptile \cite{nichol2018first} includes a similar model. Authors \cite{al2017continuous} present simple gradient-based meta-learning algorithm.

\textbf{Efficient state representations} : World models proposed in \cite{ha2018recurrent} learn a compressed spatial and temporal representation of the environment using VAEs. Further on a compact and simple policy directly from the compressed state representation. 

\subsection{Exploration issues with Imitation} 
In imitation learning, the agent makes use of trajectories provided by an expert. However, the distribution of states the expert encounters usually does not cover all the states the trained agent may encounter during testing. Furthermore imitation assumes that the actions are independent and identically distributed (i.i.d.). One solution consists in using the Data Aggregation (DAgger) methods \cite{ross2010efficient} where the end-to-end learned policy is executed, and extracted observation-action pairs are again labelled by the expert, and aggregated to the original expert observation-action dataset. Thus, iteratively collecting training examples from both reference and trained policies explores more valuable states and solves this lack of exploration. Following work on Search-based Structured Prediction (SEARN) \cite{ross2010efficient}, Stochastic Mixing Iterative Learning (SMILE) trains a stochastic stationary policy over several iterations and then makes use of a geometric stochastic mixing of the policies trained. In a standard imitation learning scenario, the demonstrator is required to cover sufficient states so as to avoid unseen states during test. This constraint is costly and requires frequent human intervention. 
More recently, Chauffeurnet \cite{bansal2018chauffeurnet} demonstrated the limits of imitation learning where even 30 million state-action samples were insufficient to learn an optimal policy that mapped bird-eye view images (states) to control (action). The authors propose the use of simulated examples which introduced perturbations, higher diversity of scenarios such as collisions and/or going off the road. The \textit{featurenet} includes an agent RNN that outputs the way point, agent box position and heading at each iteration. Authors \cite{buhet2019conditional} identify limits of imitation learning, and train a DNN end-to-end using the ego vehicles on input raw image, and 2d and 3d locations of neighboring vehicles to simultaneously predict the ego-vehicle action as well as neighbouring vehicle trajectories.


\subsection{Intrinsic Reward functions} 
In controlled simulated environments such as games, an explicit reward signal is given to the agent along with its sensor stream. However, in real-world robotics and autonomous driving deriving, designing a \textit{good} reward functions is essential so that the desired behaviour may be learned. The most common solution has been reward shaping \cite{ng1999policy} and consists in supplying additional well designed rewards to the agent to encourage the optimization into the direction of the optimal policy. Rewards as already pointed earlier in the paper, could be estimated by inverse RL (IRL) \cite{AbbeelNg2004}, which depends on expert demonstrations.
In the absence of an explicit reward shaping and expert demonstrations, agents can use intrinsic rewards or intrinsic motivation \cite{chentanez2005intrinsically} to evaluate if their actions were good or not. Authors of \cite{pathak2017curiosity} define \textit{curiosity} as the error in an agent's ability to predict the consequence of its own actions in a visual feature space learned by a self-supervised inverse dynamics model. 
In \cite{burda2018large} the agent learns a next state predictor model from its experience, and uses the error of the prediction as an intrinsic reward. This enables that agent to determine what could be a useful behavior even without extrinsic rewards.

\subsection{Incorporating safety in DRL}
Deploying an autonomous vehicle in real environments after training directly could be dangerous. Different approaches to incorporate safety into DRL algorithms are presented here. 
For imitation learning based systems, Safe DAgger \cite{SafeDAgger_AAAI2017} introduces a safety policy that learns to predict the error made by a primary policy trained initially with the supervised learning approach, without querying a reference policy. An additional safe policy takes both the partial observation of a state and a primary policy as inputs, and returns a binary label indicating whether the primary policy is likely to deviate from a reference policy without querying it.
Authors of \cite{shalev2016safe} addressed safety in multi-agent Reinforcement Learning for Autonomous Driving, where a balance is maintained between unexpected behavior of other drivers or pedestrians and not to be too defensive, so that normal traffic flow is achieved. While hard constraints are maintained to guarantee the safety of driving, the problem is decomposed into a composition of a policy for desires to enable comfort driving and trajectory planning. 
The deep reinforcement learning algorithms for control such as DDPG and safety based control are combined in \cite{xiong2016combining}, including artificial potential field method that is widely used for robot path planning. Using TORCS environment, the DDPG is applied first for learning a driving policy in a stable and familiar environment, then policy network and safety-based control are combined to avoid collisions.  It was found that combination of DRL and safety-based control performs well in most scenarios.
In order to enable DRL to escape local optima, speed up the training process and avoid danger conditions or accidents, Survival-Oriented Reinforcement Learning (SORL) model is proposed in \cite{ye2017survival}, where survival is favored over maximizing total reward through modeling the autonomous driving problem as a constrained MDP and introducing Negative-Avoidance Function to learn from previous failure. The SORL model was found to be not sensitive to reward function and can use different DRL algorithms like DDPG.
Furthermore, a comprehensive survey on safe reinforcement learning can be found in \cite{garcia2015comprehensive} for interested readers.

\subsection{Multi-agent reinforcement learning}
Autonomous driving is a fundamentally multi-agent task; as well as the ego vehicle being controlled by an agent, there will also be many other actors present in simulated and real world autonomous driving settings, such as pedestrians, cyclists and other vehicles. Therefore, the continued development of explicitly multi-agent approaches to learning to drive autonomous vehicles is an important future research direction. Several prior methods have already approached the autonomous driving problem using a MARL perspective, e.g. \cite{shalev2016safe,palanisamy2020multi,bhalla2020deep,wachi2019failure,yu2020distributed}.

One important area where MARL techniques could be very beneficial is in high-level decision making and coordination between groups of autonomous vehicles, in scenarios such as overtaking in highway scenarios \cite{yu2020distributed}, or negotiating intersections without signalised control. Another area where MARL approaches could be of benefit is in the development of adversarial agents for testing autonomous driving policies before deployment \cite{wachi2019failure}, i.e. agents controlling other vehicles in a simulation that learn to expose weaknesses in the behaviour of autonomous driving policies by acting erratically or against the rules of the road. Finally, MARL approaches could potentially have an important role to play in developing safe policies for autonomous driving \cite{shalev2016safe}, as discussed earlier.


\begin{table*}[ht]
\centering
 \begin{tabular}{l p{12cm}} \toprule
 Framework & Description \\ \midrule 
 OpenAI Baselines \cite{baselines} &  Set of high-quality implementations of different RL and DRL algorithms. The main goal for these Baselines is to make it easier for the research community to replicate, refine and create reproducible research.  \\ 
 \hline
 Unity ML Agents Toolkit \cite{juliani2018unity} &  Implements core RL algorithms, games, simulations environments for training RL or IL based agents .  \\ 
 \hline
 RL Coach \cite{caspi_itai_2017_1134899} & Intel AI Lab's implementation of modular RL algorithms implementation with simple integration of new environments by extending and reusing existing components.  \\
 \hline
 Tensorflow Agents \cite{TFAgents} & RL algorithms package with Bandits from TF. \\
 \hline
 rlpyt \cite{stooke2019rlpyt} & implements deep Q-learning, policy gradients
 algorithm families in a single python package \\
 \hline
 bsuite \cite{osband2019bsuite} & DeepMind Behaviour Suite for Reinforcement Learning aims at defining metrics for RL agents. Automating evaluation and analysis.  \\
   \bottomrule
 \end{tabular}
 \label{tab:opensource}
 \vspace{5pt}
 \caption{Open-source frameworks and packages for state of the art RL/DRL algorithms and evaluation.}
\end{table*}

\section{Conclusion} 
\label{sec:conc}

Reinforcement learning is still an active and emerging area in real-world autonomous driving applications. Although there are a few successful commercial applications, there is very little literature or large-scale public datasets available. Thus we were motivated to formalize and organize RL applications for autonomous driving. Autonomous driving scenarios involve interacting agents and require negotiation and dynamic decision making which suits RL. However, there are many challenges to be resolved in order to have mature solutions which we discuss in detail. In this work, a detailed theoretical reinforcement learning is presented, along with a comprehensive literature survey about applying RL for autonomous driving tasks.

Challenges, future research directions and opportunities are discussed in section \ref{sec:challenges}. This includes : validating the performance of RL based systems, the simulation-reality gap, sample efficiency, designing good reward functions, incorporating safety into decision making RL systems for autonomous agents. 

Reinforcement learning results are usually difficult to reproduce and are highly
sensitive to hyper-parameter choices, which are often not reported in detail. Both
researchers and practitioners need to have a reliable starting point where the well
known reinforcement learning algorithms are implemented, documented and well tested.
These frameworks have been covered in table \ref{tab:opensource}.   

The development of explicitly multi-agent reinforcement learning approaches to the autonomous driving problem is also an important future challenge that has not received a lot of attention to date. MARL techniques have the potential to make coordination and high-level decision making between groups of autonomous vehicles easier, as well as providing new opportunities for testing and validating the safety of autonomous driving policies.

Furthermore, implementation of RL algorithms is a challenging task for researchers and practitioners. This work presents examples of well known and active open-source RL frameworks that provide well documented implementations that enables the opportunity of using, evaluating and extending different RL algorithms.    
Finally, We hope that this overview paper encourages further research and applications.

\begin{table}[ht]
    \centering
    \begin{tabular}{l l} \toprule
A2C, A3C     & Advantage Actor Critic, Asynchronous A2C \\
BC      & Behavior Cloning \\
DDPG    & Deep DPG \\
DP      & Dynamic Programming \\
DPG     & Deterministic PG\\
DQN     & Deep Q-Network \\
DRL     & Deep RL \\
IL      & Imitation Learning \\
IRL     & Inverse RL \\
LfD     & Learning from Demonstration \\
MAML    & Model-Agnostic Meta-Learning \\
MARL    & Multi-Agent RL \\
MDP     & Markov Decision Process \\
MOMDP   & Multi-Objective MDP \\
MOSG    & Multi-Objective SG \\
PG      & Policy Gradient \\
POMDP   & Partially Observed MDP \\
PPO     & Proximal Policy Optimization \\
QL      & Q-Learning \\
RRT     & Rapidly-exploring Random Trees \\
SG      & Stochastic Game \\
SMDP    & Semi-Markov Decision Process \\
TDL     & Time Difference Learning \\
TRPO    & Trust Region Policy Optimization \\ \bottomrule
    \end{tabular}
    \vspace{10pt}
    \caption{Acronyms related to Reinforcement learning (RL).}
    \label{tab:acronym1}
\end{table}


\bibliographystyle{IEEEtran}
\bibliography{bibliography}

\begin{IEEEbiography}
[{\includegraphics[width=1in,height=1.25in,clip,keepaspectratio]{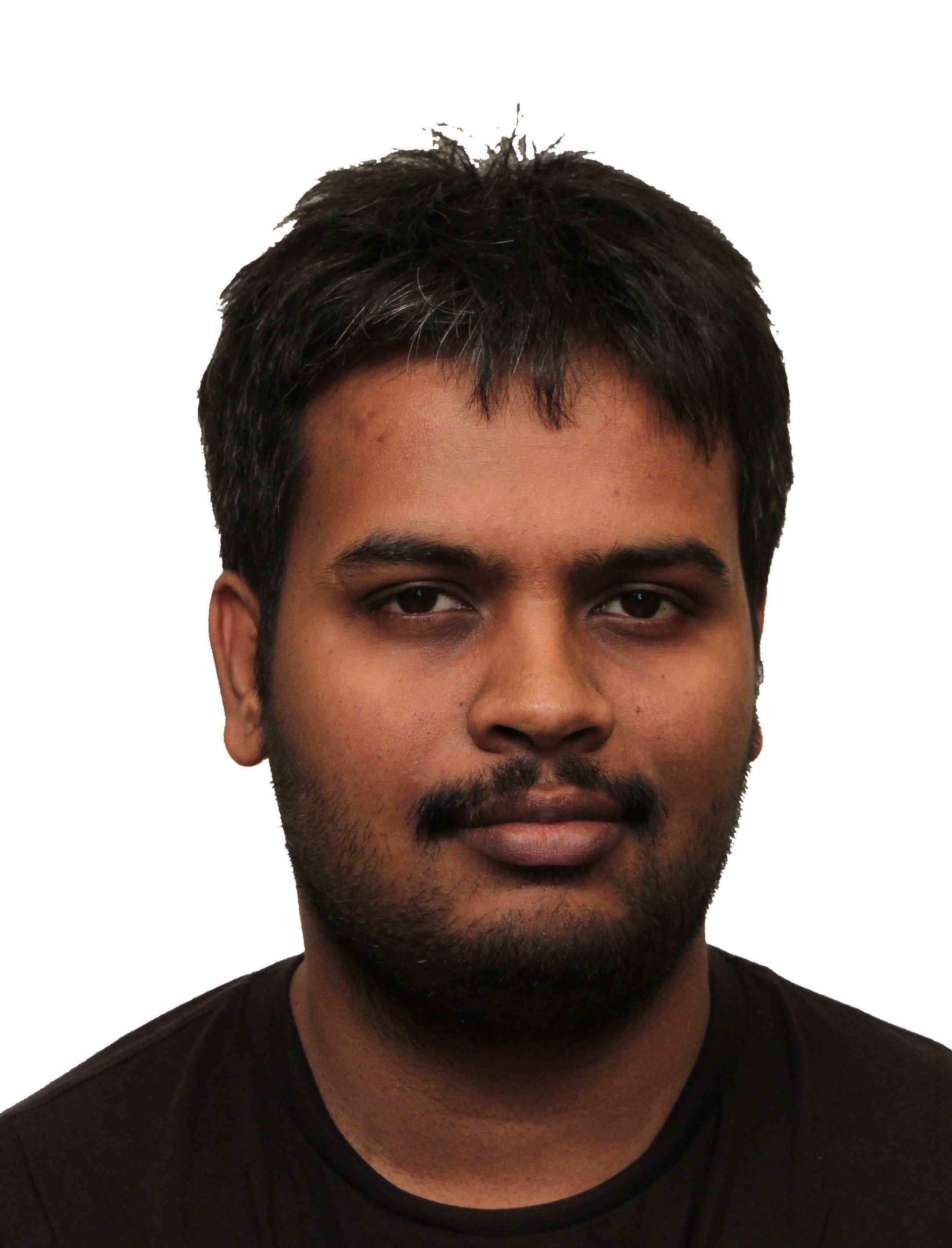}}]{B Ravi Kiran}
is the technical lead of machine learning team at Navya, designing and deploying realtime deep learning architectures for perception tasks on autonomous shuttles. During his 6 years in academic research, he has worked on DNNs for video anomaly detection, online time series anomaly detection, hyperspectral image processing for tumor detection. He finished his PhD at Paris-Est in 2014 entitled Energetic lattice based optimization which was awarded the MSTIC prize. He has worked in academic research for over 4 years in embedded programming and in autonomous driving. During his career he has published over 40 articles and journals.
\end{IEEEbiography}
\vspace{-1cm}
\begin{IEEEbiography}
[{\includegraphics[width=1in,height=1.25in,clip,keepaspectratio]{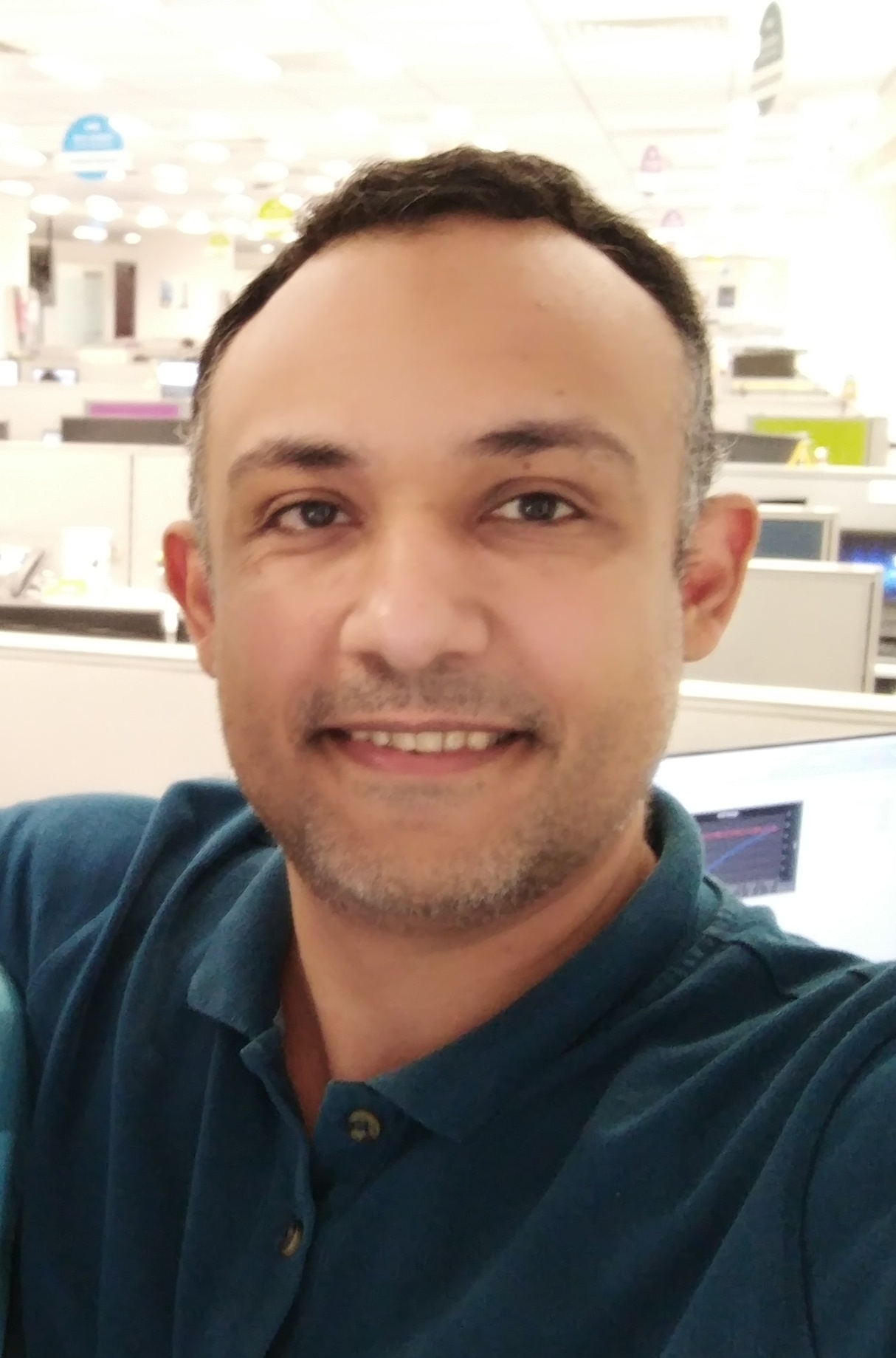}}]{Ibrahim Sobh}
Ibrahim has more than 20 years of experience in the area of Machine Learning and Software Development. Dr. Sobh received his PhD in Deep Reinforcement Learning for fast learning agents acting in 3D environments. He received his B.Sc. and M.Sc. degrees in Computer Engineering from Cairo University Faculty of Engineering. His M.Sc. Thesis is in the field of Machine Learning applied on automatic documents summarization. Ibrahim has participated in several related national and international mega projects, conferences and summits. He delivers training and lectures for academic and industrial entities. Ibrahim's publications including international journals and conference papers are mainly in the machine and deep learning fields. His area of research is mainly in Computer vision, Natural language processing and Speech processing. Currently, Dr. Sobh is a Senior Expert of AI, Valeo.
\end{IEEEbiography}
\vspace{-1cm}
\begin{IEEEbiography}
[{\includegraphics[width=1in,height=1.25in,clip,keepaspectratio]{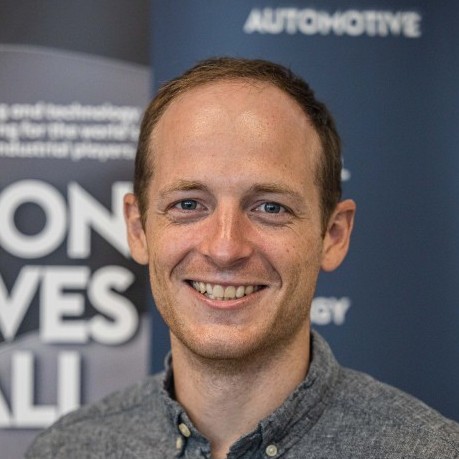}}]{Victor Talpaert}
is a PhD student at U2IS, ENSTA Paris, Institut Polytechnique de Paris, 91120 Palaiseau, France. His PhD is directed by Bruno Monsuez since 2017, the lab speciality is in robotics and complex systems. His PhD is co-sponsored by AKKA Technologies in Gyuancourt, France, through the guidance of AKKA's Autonomous Systems Team. This team has a large focus on Autonomous Driving (AD) and the automotive industry in general. His PhD subject is learning decision making for AD, with assumptions such as a modular AD pipeline, learned features compatible with classic robotic approaches and ontology based hierarchical abstractions.
\end{IEEEbiography}
\vspace{-1cm}
\begin{IEEEbiography}
[{\includegraphics[width=1in,height=1.25in,clip,keepaspectratio]{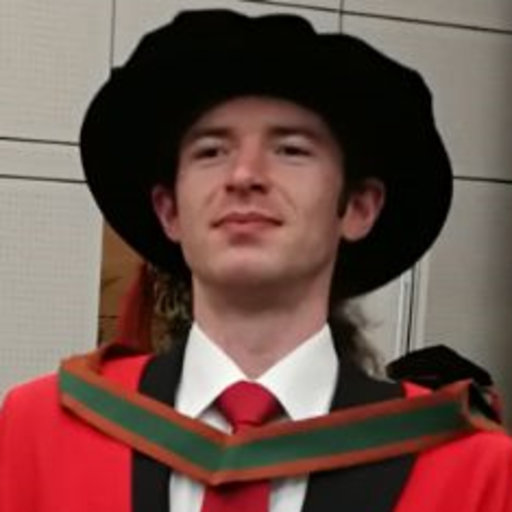}}]{Patrick Mannion}
is a permanent member of academic staff at National University of Ireland Galway, where he lectures in Computer Science. He is also Deputy Editor of The Knowledge Engineering Review journal. He received a BEng in Civil Engineering, a HDip in Software Development and a PhD in Machine Learning from National University of Ireland Galway, a PgCert in Teaching \& Learning from Galway-Mayo IT and a PgCert in Sensors for Autonomous Vehicles from IT Sligo. He is a former Irish Research Council Scholar and a former Fulbright Scholar. His main research interests include machine learning, multi-agent systems, multi-objective optimisation, game theory and metaheuristic algorithms, with applications to domains such as transportation, autonomous vehicles, energy systems and smart grids.
\end{IEEEbiography}
\vspace{-1cm}
\begin{IEEEbiography}
[{\includegraphics[width=1in,height=1.25in,clip,keepaspectratio]{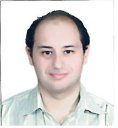}}]{Ahmad El Sallab}
Ahmad El Sallab is the Senior Chief Engineer of Deep Learning at Valeo Egypt, and Senior Expert at Valeo Group. Ahmad has 15 years of experience in Machine Learning and Deep Learning, where he acquired his M.Sc. and Ph.D. on 2009 and 2013 in the field. He has worked for reputable multi-national organizations in the industry since 2005 like Intel and Valeo. He has over 35 publications and book chapters in Deep Learning in top IEEE and ACM journals and conferences, in addition to 30 patents, with applications in Speech, NLP, Computer Vision and Robotics.
\end{IEEEbiography}
\vspace{-1cm}
\begin{IEEEbiography}
[{\includegraphics[width=1in,height=1.25in,clip,keepaspectratio]{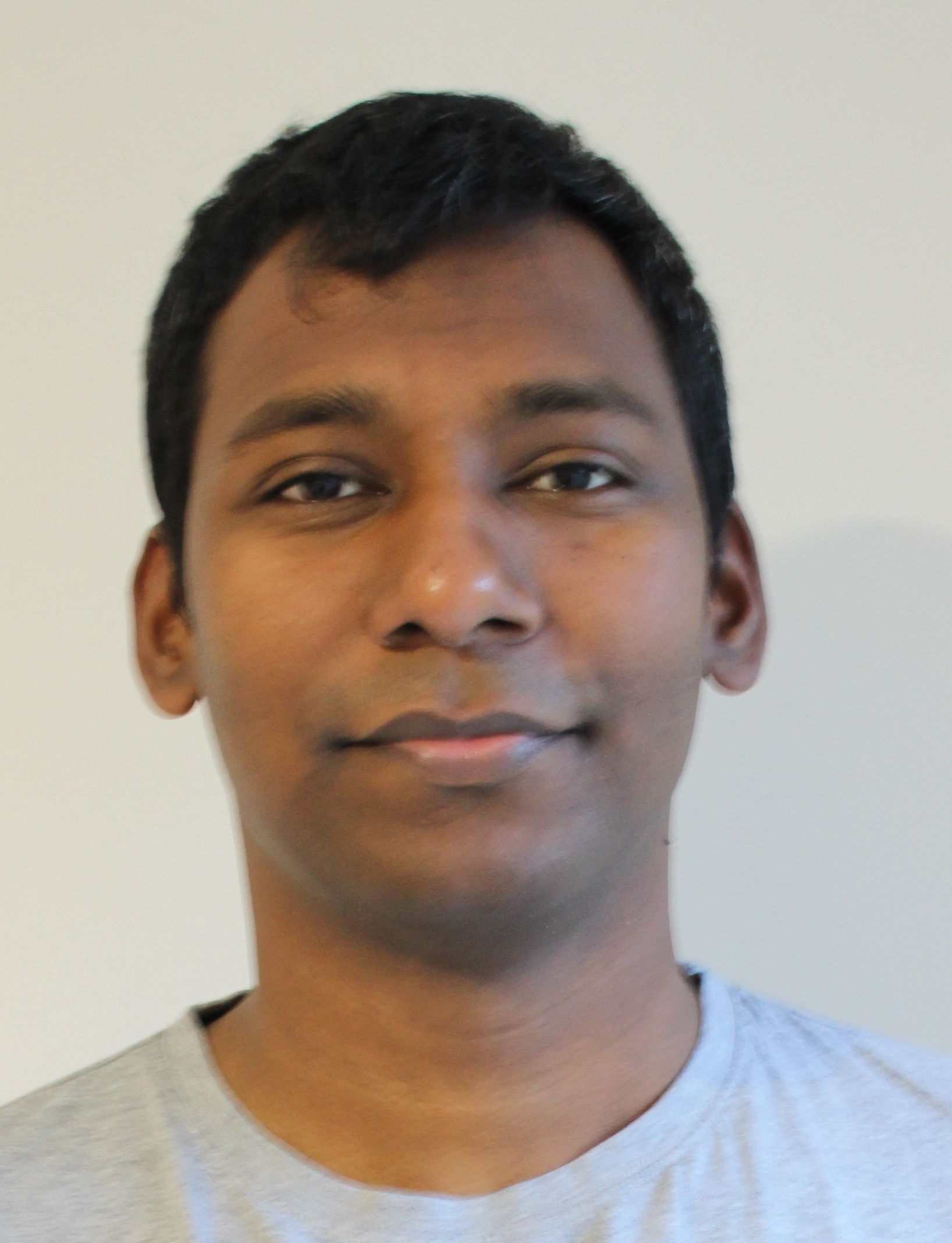}}]{Senthil Yogamani} is an Artificial Intelligence architect and technical leader at Valeo Ireland. He leads the research and design of AI algorithms for various modules of autonomous driving systems. He has over 14 years of experience in computer vision and machine learning including 12 years of experience in industrial automotive systems. He is an author of over 90 publications and 60 patents with 1300+ citations. He serves in the editorial board of various leading IEEE automotive conferences including ITSC, IV and ICVES and advisory board of various industry consortia including Khronos, Cognitive Vehicles and IS Auto. He is a recipient of best associate editor award at ITSC 2015 and best paper award at ITST 2012.
\end{IEEEbiography}
\vspace{-1cm}
\begin{IEEEbiography}
[{\includegraphics[width=1in,height=1.25in,clip,keepaspectratio]{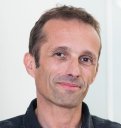}}]{Patrick P\'erez}
 is Scientific Director of Valeo.ai, a Valeo research lab on artificial intelligence for automotive applications. He is currently on the Editorial Board of the International Journal of Computer Vision. Before joining Valeo, Patrick P\'erez has been Distinguished Scientist at Technicolor (2009-2918), researcher at Inria (1993-2000, 2004-2009) and at Microsoft Research Cambridge (2000-2004). His research interests include multimodal scene understanding and computational imaging. 
\end{IEEEbiography}

\end{document}